\documentclass[runningheads]{llncs}
\usepackage[T1]{fontenc}
\usepackage{graphicx}
\usepackage{xcolor}
\usepackage{kotex}

\begin{document}
\title{A Visuo-Tactile Data Collection System\\with Haptic Feedback for Coarse-to-Fine Imitation Learning}
%
\titlerunning{Visuo-Tactile Data Collection System}
%
\author{
Yeseung Kim$^{*}$ \and
Nayoung Oh$^{*}$ \and 
Jun Park$^{*}$ \and 
Teetat Thamronglak$^{*}$ \and
Daehyung Park$^{**}$
}
\authorrunning{Y. Kim et al.}
%
\institute{Korea Advanced Institute of Science and Technology, Yuseong-gu Daehak-ro 291, Daejeon 34141, South Korea
}
\maketitle              

\def\thefootnote{*}\footnotetext{These authors contributed equally to this work}\def\thefootnote{\arabic{footnote}}
\def\thefootnote{**}\footnotetext{Corresponding author: \email{daehyung@kaist.ac.kr}}\def\thefootnote{\arabic{footnote}}

\begin{abstract}
We present a visuo-tactile data-collection system that generates temporally structured, contact-rich demonstrations for imitation learning. Conventional systems often decouple the operator from contact forces, which hinders the demonstration of subtle force modulation. Our system introduces a direct-drive gripper that the operator actuates with the fingers, preserving natural haptic feedback. Integrated visual sensors and custom tactile arrays capture image streams and contact geometry. A handle-mounted push button enables the operator to annotate the task’s temporal structure in real time by marking task-critical regions. By fusing in-hand force perception with in-situ temporal annotation, the system produces multimodal datasets designed for coarse-to-fine learning algorithms that exploit structural task knowledge, enabling the development of high-quality manipulation policies.

\end{abstract}
\section{Introduction}
Imitation learning (IL) enables robots to acquire dexterous, generalizable skills directly from human demonstrations---a particularly effective paradigm for contact-rich manipulation where precision and robustness are essential. To enhance the quality of IL policies, demonstration datasets need to satisfy two criteria: (i) high-quality human demonstrations collected through physical interaction with the environment; and (ii) task knowledge that encodes temporal structure---coarse approach motions to reach the task bottleneck, followed by fine adjustments to complete the interaction~\cite{Zhao-RSS-23,li-iros23,coarse-to-fine,dispo}. 

Conventional data collection pipelines~\cite{diffusion-policy,umi,viola} fail to provide such data. First, kinematic interfaces decouple the operator from contact forces, making it difficult to feel the feedback from the interaction and preventing subtle force modulation essential for fine control~\cite{mimictouch,zhang-rss25}. 
Second, there is no built-in mechanism for marking task-critical regions or transitions (e.g., the switch from coarse approach to fine adjustment), so datasets lack the structural labels expected by coarse-to-fine methods~\cite{coarse-to-fine,dispo} 

To address these limitations, we propose a novel data collection device designed for temporal structure-aware, contact-rich demonstrations. The operator actuates the gripper jaws directly with their fingers, preserving natural haptic feedback, while integrated sensors record synchronized visuo-tactile signals during contact~\cite{3d-vitac,umi}. In addition, a handle-mounted button allows the operator to label critical regions in real time. These labels are time-aligned with the sensor streams, making the collected demonstrations usable by coarse-to-fine learning algorithms that rely on phase or region annotations~\cite{coarse-to-fine,dispo}.
\section{Related Work}
\subsection{Tactile Sensing for Robotic Manipulation}
Effective manipulation in contact-rich environments relies on tactile feedback~\cite{park2014interleaving,mao-natc24,funk-arxiv25}, which conveys force, texture, and shape information that vision alone cannot capture. Vision-based tactile sensors~\cite{agarwal-comm25,guzey-corl23}, in particular, offer high-resolution, information-rich signals. Our work utilizes the $3$D-ViTac sensor~\cite{3d-vitac}, a vision-based tactile modality capable of reconstructing detailed $3$D contact geometry. However, a key challenge remains: collecting expert demonstrations that make full use of this tactile information while preserving the operator’s natural haptic perception.

\subsection{Data Collection for Imitation Learning}
Data collection device is the prevailing paradigm for collecting expert demonstrations for IL~\cite{wang-corl23,wong-corl21,brown-corl19,mandlekar-corl18}. Devices such as the UMI gripper~\cite{umi} streamline this process by providing high-fidelity $6$D pose tracking of the gripper while it is hand-held by a human operator. However, interfaces like pistol grips physically decouple the operator from the gripper’s end-effector, preventing direct perception of contact forces. This, in turn, makes it difficult to demonstrate the force-sensitive actions required for fine-grained manipulation~\cite{mimictouch}.

\subsection{Coarse-to-Fine Imitation Learning}
Traditional IL approaches often learn a monolithic policy for an entire task~\cite{diffusion-policy,viola,zhang-rss24}, which can be inefficient when high precision is applied to simple, non-critical movements. Recent research therefore adopts coarse-to-fine frameworks. Coarse-to-Fine Imitation Learning models a manipulation task as two sequential phases: a coarse free-space approach that brings the end-effector to a task-specific pose followed by a fine interaction phase executed by repeating the demonstration’s end-effector velocities~\cite{coarse-to-fine}. DiSPo adjusts action granularity dynamically, generating coarse motions in less critical regions and fine-grained control in critical areas to improve efficiency and performance~\cite{dispo}. A common requirement across these methods is data with annotations that reflect the underlying task structure. Conventional data collection, however, typically lacks intuitive mechanisms for providing such labels, motivating the design of our data collection device.

\begin{figure}[t]
\centering
\includegraphics[width=\textwidth]{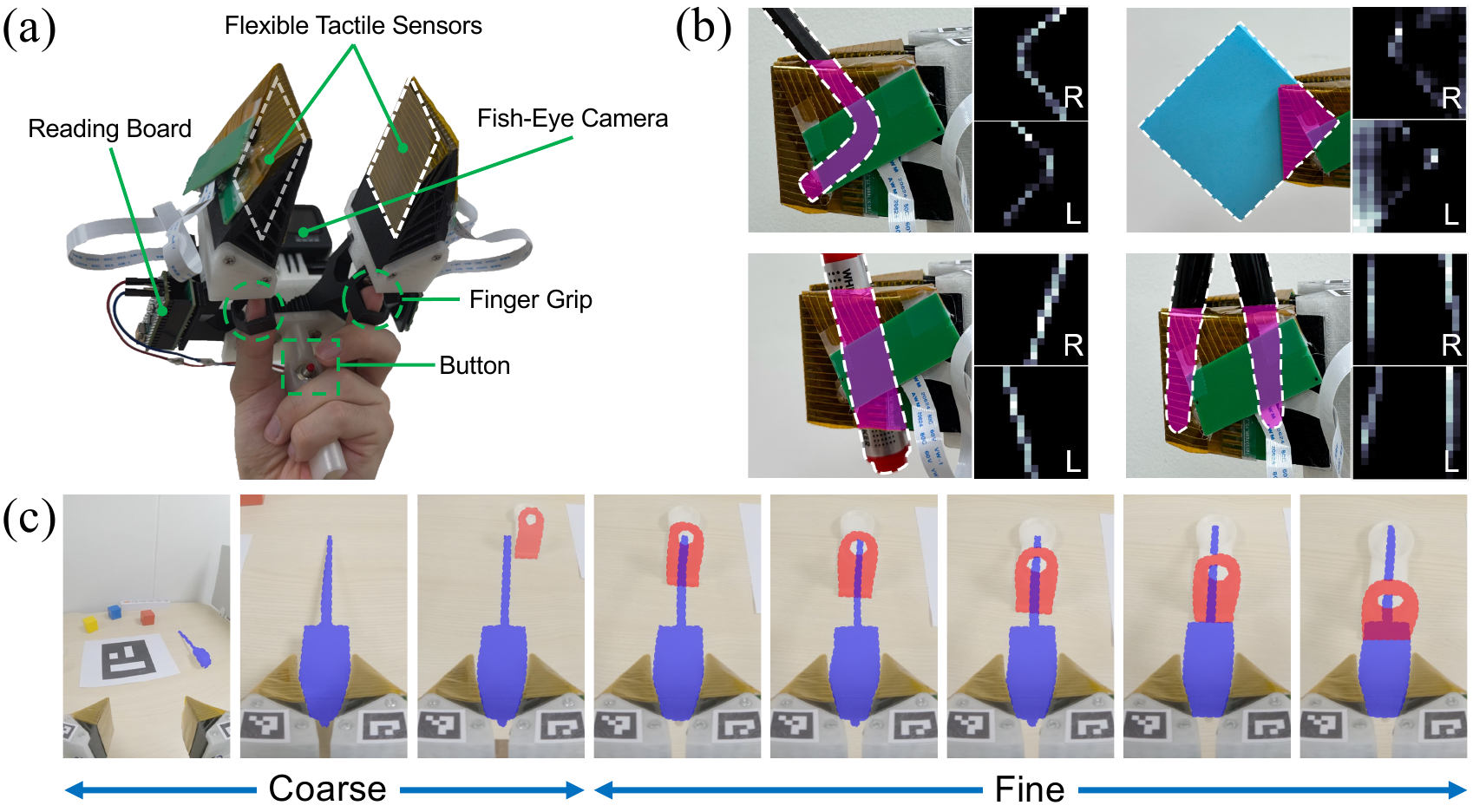}
\caption{Overview of our visuo-tactile data collection system. (a) The hardware design, featuring direct-drive finger grips for haptic feedback, tactile sensors, a camera, and an annotation button. (b) Sample tactile readings from the left (L) and right (R) sensors during contact with different objects. The white boundary outlines the entire object, while the magenta overlay indicates the part occluded by the gripper. (c) An example image sequence of a demonstration captured by the attached camera. We crop the images for better visualization and highlight the task objects in red and purple colors. We use the annotation button to label each segment as coarse or fine; we then visualize the resulting segments.}
\label{fig:system_overview}
\end{figure}

\section{Proposed System}
We propose a visuo–tactile data collection system that collects rich, temporally structured demonstrations tailored for coarse-to-fine imitation learning. Building on the design principles of portable interfaces such as UMI~\cite{umi}---which use an ego-centric camera to reduce embodiment mismatch with a robot---our system adds two capabilities aligned with the introduction: (i) preserving natural haptic perception by allowing the operator to actuate the gripper directly, and (ii) enabling in-situ annotation of task-critical regions so that the temporal structure of a task is explicitly captured. The following sections detail the hardware and the data collection process.

\subsection{Hardware Design}

As shown in Fig.~\ref{fig:system_overview}(a), the system integrates three components on a custom $3$D-printed gripper base: a visual sensor, tactile sensors, and an annotation interface.

\textbf{Gripper.}
The end-effector is a two-jaw gripper whose jaws are actuated directly by the operator’s fingers via a compact mechanical linkage. This arrangement maintains a clear mapping between human finger motion and gripper width, allowing the operator to perceive contact-induced resistance and modulate force naturally. The tactile modules are mounted directly on the inner surfaces of the jaws so that measurements are captured exactly at the contact interface.

\textbf{Visual Sensor.}
For robust visual tracking, we mount a GoPro camera with a fisheye lens above the gripper. Following UMI~\cite{umi}, the wide field of view preserves sufficient context for tracking even under partial self-occlusion. Unlike the original UMI design, we do not employ side mirrors. 
Fig.~\ref{fig:system_overview}(c) illustrates an example demonstration image sequence captured by the attached camera.

\textbf{Tactile Sensor.}
To provide direct force-related feedback, we integrate custom $3$D-ViTac sensors~\cite{3d-vitac} onto the left and right jaws. Each pad adopts a triple-layer construction: a piezoresistive sheet between two orthogonal sets of conductive threads, forming a dense $16\times16$ array per jaw. This configuration yields detailed measurements of contact presence and normal force and supports reconstruction of contact geometry. In particular, the visuo–tactile signals at the jaw–object interface allow us to recover local surface shape even in regions that are occluded by the gripper in the camera view. Fig.~\ref{fig:system_overview}(b) shows examples of the tactile data captured by the sensors when grasping different objects.

\textbf{Annotation Interface.}
A push button on the handle allows the operator to mark task-critical regions or transitions in real time. An Arduino microcontroller reads both the tactile arrays and the button state and records time-stamped signals for temporal alignment. In Fig.~\ref{fig:system_overview}(c), we visualize the button-derived labels: the operator holds the button to label fine segments and releases it to label coarse segments.

\subsection{Data Collection and Processing}

We design our data collection pipeline to produce synchronized, multi-modal streams suitable for policy learning. The process begins with the simultaneous collection of video from the ego-centric camera and tactile data from the 3D-ViTac sensors as the operator performs a demonstration. We record data from all sensors, including the annotation button, with precise timestamps. Following the demonstration, we estimate the $6$-DoF pose of the end-effector in a post-processing step. We apply the ORB-SLAM$3$~\cite{orb-slam3} algorithm to the recorded video, which localizes the gripper and computes its trajectory. We track gripper width separately using ArUco markers affixed to the jaws. Finally, we synchronize all individual data streams---visual (RGB images), proprioceptive (6-DoF pose and gripper width), tactile measurements ($16 \times 16$ force arrays), and annotation (binary signal)---using their saved timestamps. This process yields a coherent, time-indexed dataset where all modalities are precisely aligned.

\begin{figure}[t]
\centering
\includegraphics[width=\textwidth]{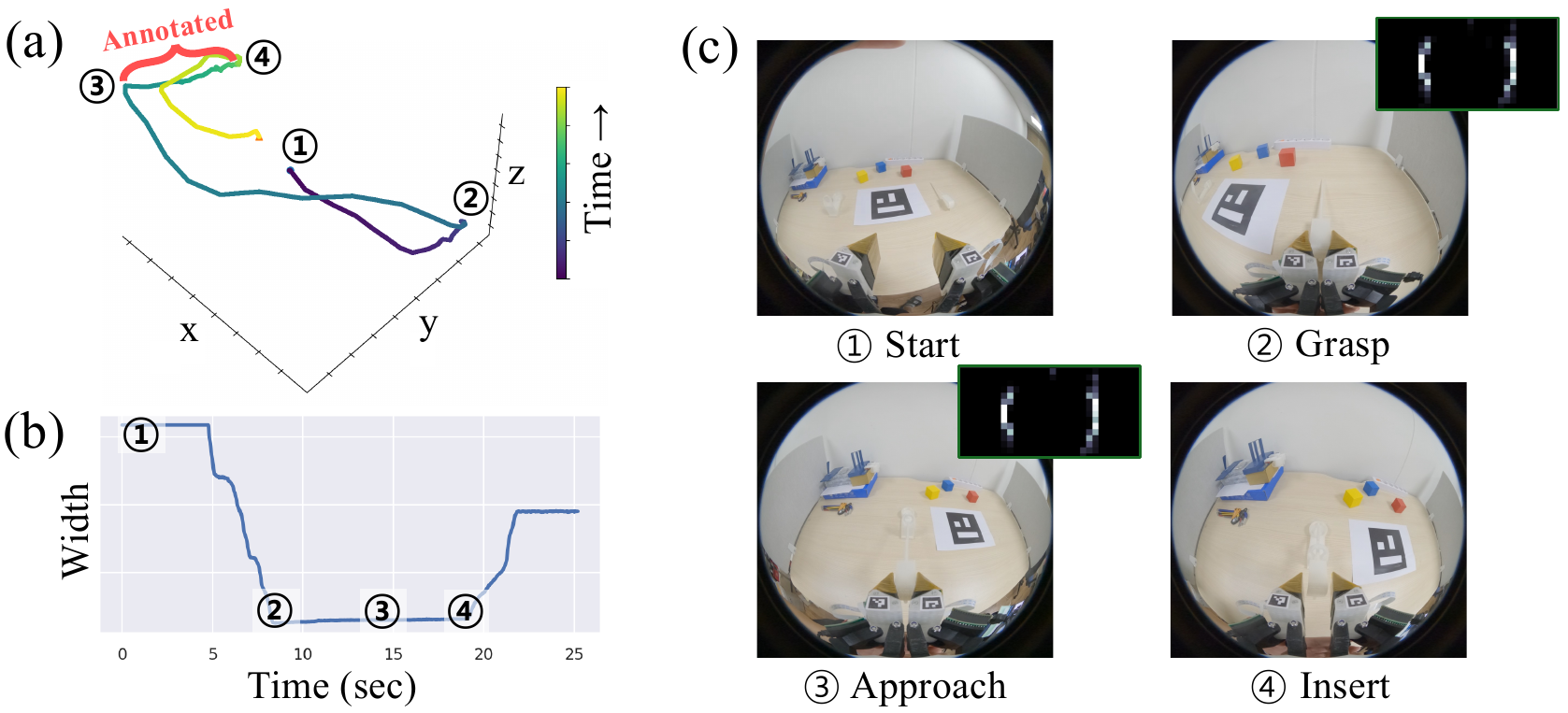}
\caption{Single-demonstration summary of the toy insertion task collected with our device: (a) gripper position trajectory (b) gripper width over time (c) images and tactile measurements of event frames.}
\label{fig:data_rep}
\end{figure}

\subsection{Data Representation}
The collected dataset consists of time-aligned sequences of visual observations (i.e., images), $6$-DoF end-effector poses, gripper widths, tactile readings, and critical-region labels. 
Fig.~\ref{fig:system_overview}(c) and Fig.~\ref{fig:data_rep} illustrates a single demonstration collected with our device on a toy insertion task: grasping a hex wrench equipped with a grasp-assisting attachment and inserting it into the hole of a tilted stand. 
In Fig.~\ref{fig:system_overview}(c), we depict the task sequence segmented into two phases: a coarse phase covering the approach and grasp, and a fine phase covering precision alignment and insertion.
Since the stand is tilted at an angle, subtle alignment is important yet difficult to infer reliably from images alone. Consequently, the shape of the tactile imprint provides a direct signal for the correct insertion angle. In Fig.~\ref{fig:data_rep}(a), we plot the gripper’s 3D position trajectory estimated via SLAM algorithm; the interval highlighted in red is annotated as critical due to the tight clearance at the hole. 
We mark important events as numbered markers: (1) initial pose, (2) wrench grasp, (3) approach to the stand hole, and (4) completion of insertion. 
Fig.~\ref{fig:data_rep}(b) reports the gripper width over time, showing closure at event (2) and reopening after event (4). Fig.~\ref{fig:data_rep}(c) presents camera frames at each event; the tactile measurements exhibit the wrench imprint during events (2) and (3).

\section{Conclusion}
In this paper, we introduce a novel visuo-tactile data collection system to address key limitations in conventional data collection methods for imitation learning. By enabling operators to actuate the gripper directly, our system preserves the natural haptic feedback crucial for demonstrating contact-rich manipulation tasks. Furthermore, the integration of a real-time annotation mechanism allows for the explicit capture of a task's temporal structure, providing critical labels for coarse-to-fine learning algorithms. 
The resulting datasets---rich with synchronized visual and tactile streams, gripper $6$-DoF pose trajectories, and gripper-width measurements---have a structure that facilitates the development of policies that distinguish between broad approach motions and fine, contact-heavy interactions. We believe this system provides a valuable tool for advancing research in robotic manipulation, particularly for tasks such as inverse constraint learning that recovers constraints from demonstrations~\cite{park2020inferring,jang2023inverse,cho2025ilcl}, and large language model–assisted imitation learning~\cite{kim2024survey}.

\subsubsection*{Acknowledgements.}
This research was supported by the MSIT (Ministry of Science and ICT), Korea, under the ITRC (Information Technology Research Center) support program (IITP-2025-RS-2024-00437102) supervised by the IITP (Institute for Information \& Communications Technology Planning \& Evaluation), the National Research Council of Science \& Technology(NST) grant by the Korea government(MSIT) (No. GTL25041-000), Artificial intelligence industrial convergence cluster development project funded by the MSIT \& Gwangju Metropolitan City, and the KAIST Convergence Research Institute Operation Program.

\bibliographystyle{splncs04}
\bibliography{references}

\end{document}